%% file: main.tex
\documentclass[10pt,twocolumn,letterpaper]{article}

\usepackage[pagenumbers]{cvpr}      

\usepackage[ruled]{algorithm2e} 
\usepackage{soul} 

\usepackage{booktabs} 
\usepackage{graphicx}
\usepackage{amsmath}
\usepackage{amssymb}
\usepackage{booktabs}
\usepackage{multirow}
\usepackage{commath} 
\usepackage{float}
\usepackage{pifont} 

\SetAlFnt{\small}
\SetAlCapFnt{\small}
\SetAlCapNameFnt{\small}
\SetAlCapHSkip{0pt}
\usepackage{amsmath}

\usepackage[pagebackref,breaklinks,colorlinks]{hyperref}

\makeatletter
\@namedef{ver@everyshi.sty}{}
\makeatother
 \usepackage{tikz,pgfplots}
\usetikzlibrary{positioning}
\usepackage{arydshln}

\usepackage[capitalize]{cleveref}
\crefname{section}{Sec.}{Secs.}
\Crefname{section}{Section}{Sections}
\Crefname{table}{Table}{Tables}
\crefname{table}{Tab.}{Tabs.}

\definecolor{darkg}{rgb}{0,0.4,0}
\definecolor{lgreen}{rgb}{0,0.6,0}
\definecolor{amethyst}{rgb}{0.6, 0.4, 0.8}
\definecolor{orange}{rgb}{0.93,0.48,0.03}
\definecolor{blueviolet}{rgb}{0.54,0.16,0.88}
\definecolor{purple}{rgb}{0.5,0,0.5}

\newcommand{\afterfigure}{\vspace*{-0.2cm}}
\newcommand{\shortcite}[1]{\cite{#1}}

\usepackage[margin={1.2cm,2.2cm}]{geometry}
\usepackage{multicol}

\begin{document}

\title{ Self-Distilled StyleGAN: Towards Generation from Internet Photos}


\author{Ron Mokady$^{1,2}\thanks{Performed this work while an intern at Google.}$
\qquad
Michal Yarom$^{1}$
\qquad
Omer Tov$^{1}$
\qquad
Oran Lang$^{1}$
\\
\qquad
Daniel Cohen-Or$^{2}$
\qquad
Tali Dekel$^{1,3}$
\qquad
Michal Irani$^{1,3}$
\qquad
Inbar Mosseri$^{1}$
\\
\small $^1$  Google Research \qquad $^2$ Tel Aviv University\qquad  $^3$ Weizmann Institute of Science
}

\maketitle

\begin{abstract}
\input{abstract}

\end{abstract}

\begin{figure}[h]
\centering
\vspace{-0.1cm}
\includegraphics[width=\columnwidth]{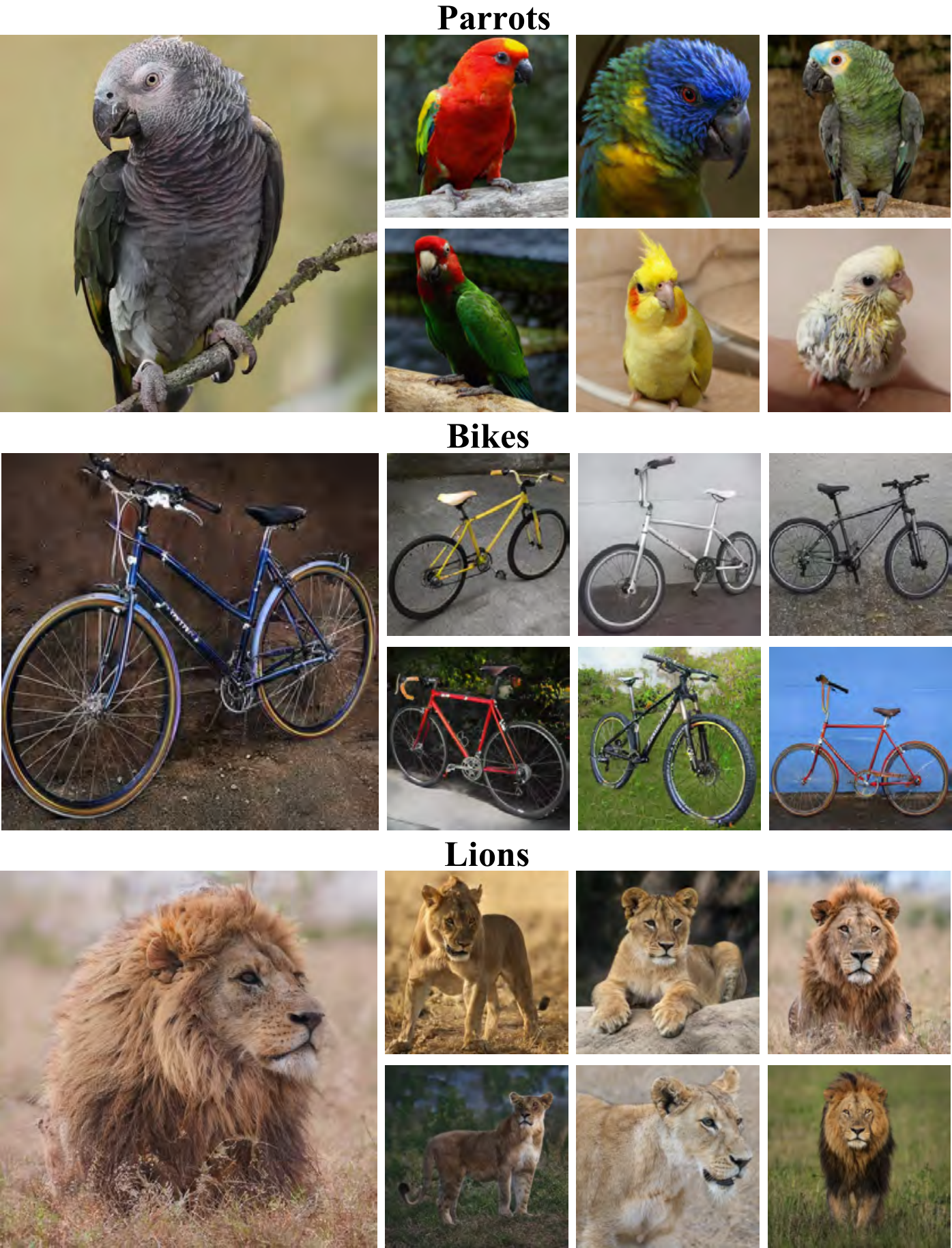}
 \vspace{-0.5cm}
\caption{  {{\textbf{Our Self-Distilled StyleGAN framework enables high-quality and diverse image generation from uncurated Internet datasets.}}} {\it Self-distillation is performed in two manners: (i) a dedicated self-filtering approach, which distills the raw noisy data by removing outlier and undesirable images, and (ii) multi-modal ``truncation trick'' based on perceptual clustering in StyleGAN's latent space, which allows to reduce visual artifacts while preserving better the diversity of the samples images. }}
\label{fig:teaser}
\afterfigure
\end{figure}

\input{intro}

\input{related}

\input{method}

\input{results}

\input{conclusion}

\newpage

\bibliographystyle{ieee_fullname}
\bibliography{egbib}
\newpage

\appendix

\input{appendix}

\end{document}

%% file: abstract.tex

StyleGAN is known to produce high-fidelity images, while also offering unprecedented semantic editing. However, these fascinating abilities have been demonstrated only on a limited set of datasets, which are usually structurally aligned and well curated. In this paper, we show how StyleGAN can be adapted to work on raw uncurated images collected from the Internet. Such image collections impose two main challenges to StyleGAN: they contain many outlier images, and are characterized by a multi-modal distribution. Training StyleGAN on such raw image collections results in degraded image synthesis quality. To meet these challenges, we proposed a \emph{StyleGAN-based self-distillation} approach, which consists of two main components: (i)~A generative-based self-filtering of the dataset to eliminate outlier images, in order to generate an adequate training set, and (ii)~Perceptual clustering of the generated images to detect the inherent data modalities, which are then employed to improve StyleGAN's ``truncation trick'' in the image synthesis process. The presented technique enables the generation of high-quality images, while minimizing the loss in diversity of the data. Through qualitative and quantitative evaluation, we demonstrate the power of our approach to new challenging and diverse domains collected from the Internet. New datasets and pre-trained models are available in our project website \footnote{\bf Project website: \ \scalebox{0.91}[1.0]{\bf \url{https://self-distilled-stylegan.github.io/}}}.

%% file: intro.tex
\section{Introduction}

\input{figures/architecture}

StyleGAN~\cite{karras2019style, karras2020analyzing} has recently been established as one of the most prominent generative models.
Not only does it achieve phenomenal visual quality, but it also demonstrates breathtaking semantic editing capabilities due to its highly disentangled latent space. Various applications have exploited these traits in the human facial domain, such as the editing of the pose, age, and expression. However, the synthesis abilities of StyleGAN were extended to only a handful of other domains, which are highly structurally aligned and curated.
Since such manually curated datasets are rare, employing StyleGAN over different domains is rather limited.

A natural source for large image datasets is the Internet. However, image collections crawled from the Internet impose two main challenges to StyleGAN: 
First, the data gathered from the Internet is noisy, containing many outliers or irrelevant images. For instance, when gathering images of a ``dog'', we may get images containing cartoons, sculptures, or even images portraying a dog with out-of-domain objects, such as a cat standing next to a dog.
 Second, the image distribution of these datasets is multi-modal. Unlike the rather aligned facial dataset, most domains are highly diverse in terms of geometry and texture. For example, images of full-body dogs not only depict various breeds but also exhibit a wide range of poses and camera settings. 
 Consequently, Training StyleGAN on such raw uncurated image collections results in degraded image synthesis quality.
 Generally speaking, there is a natural tradeoff between the image quality and the diversity of the unconditionally generated images~\cite{luvcic2019high, liu2020diverse}.

    In this work, we tackle the challenges posed by Internet image collections by proposing a StyleGAN-based self-distillation approach. This approach consists of two main components. 
    Prior to the training procedure, we perform an \emph{unsupervised} generative-based self-filtering, where outliers are omitted.  
    Our key idea is to use a \textit{self-prior}, where the generator itself conducts the filtering process. Our approach is based on the intrinsic bias of the generator to focus on objects with high prevalence, rather than eccentricities. For instance, when training over elephant photos, the generator struggles to portray occasional humans.
    By first training StyleGAN over the uncurated collection, we can automatically filter out images according to their reconstruction quality attainable by this generator. We set a \emph{dataset-specific filtering threshold} 
   in a way which guarantees a reasonable trade-off between the \emph{diversity} of the filtered images vs. their \emph{reconstruction quality}.
    We show that re-training StyleGAN over the filtered subset produces high-quality and diverse images.

    Nevertheless, the filtered collection is still quite complex, spanning a wide range of poses, zooms, and appearance variations. To handle this,
   we introduce a post-training \emph{multi-modal variant} of StyleGAN's commonly used ``truncation trick'' on the latent code~\cite{karras2019style, karras2020analyzing}. Since StyleGAN generates a latent code from a Gaussian noise vector, some generated codes represent the distribution margins, resulting in flawed and erroneous synthesis. Hence, the generated latent codes are often interpolated with the global mean latent code, in an operation referred to as the \textit{truncation trick}  for StyleGAN. However, in more challenging  domains,  that are characterized by a multi-modal distribution, such interpolation with the single mean vector incurs a mode collapse, thus resulting in inferior diversity of the generated images. For example, in a very colorful Parrots collection (see Fig.~\ref{fig:teaser}),
   the mean latent vector results in a green average parrot (Fig.~\ref{fig:truncation_single_image}b). Truncating new generated latent codes to this global mean induces greenish feathers in all generated samples  (Figures~\ref{fig:truncation_grid}b and~\ref{fig:truncation_single_image}c).
   Instead, we propose to first \emph{identify and cluster} different data modalities within the space of latent codes of StyleGAN. We then assign each new generated latent code to its ``nearest'' cluster center, and truncate it toward that cluster center (instead of the global mean). We show that by applying such multi-modal truncation, we maintain a high-quality generation while keeping the diversity of the multi-modal distribution.

    Without using any supervision, our self-distilled StyleGAN can successfully handle much more complicated and diverse datasets than the currently used ones. In particular, we create numerous diverse and multi-modal filtered datasets: Dogs, Horses, Bicycles, Lions, Elephants, Potted plants and Parrots. For instance, the Lions dataset contains different body poses and camera views 
    (see Fig.~\ref{fig:teaser}). 
    Through extensive qualitative and quantitative results, we validate that our approach substantially outperforms native StyleGAN training over un-curated web image collections. Furthermore, we show that the realistic editing capabilities of StyleGAN remain applicable for these new challenging datasets. Additional visual results are provided in the \href{https://self-distilled-stylegan.github.io/supplementary/index.html}{project website results page}.

%% file: figures/architecture.tex
   \begin{figure*}
 {\includegraphics[width=\textwidth]{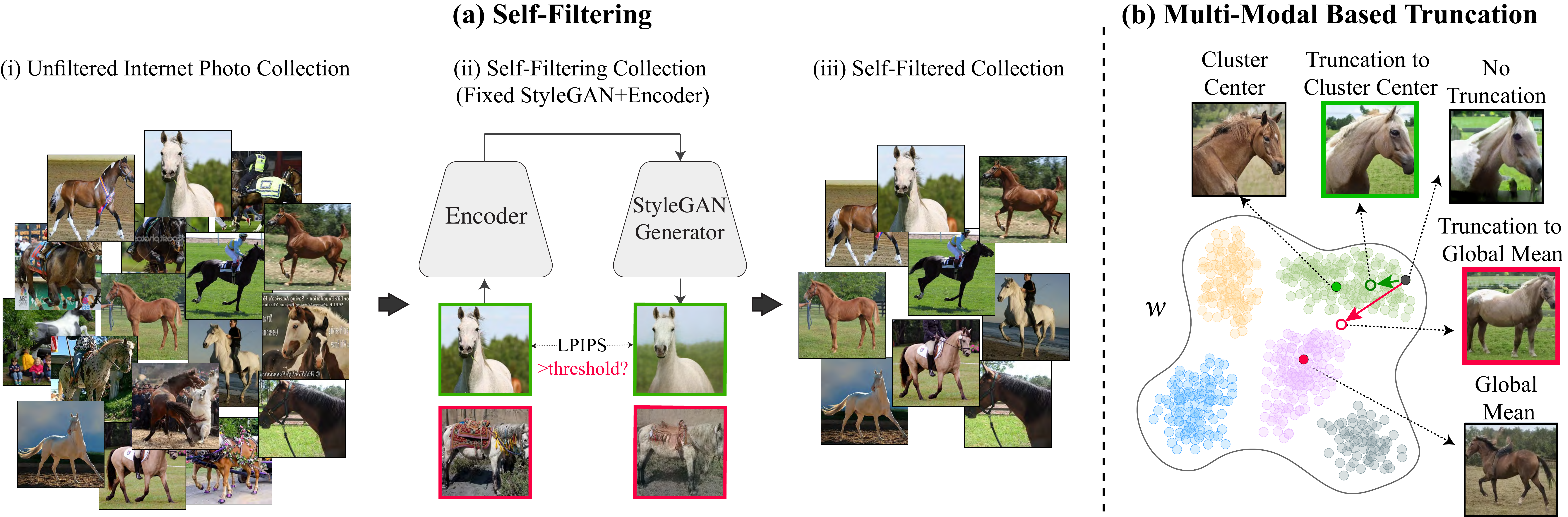}}
\vspace*{-0.2cm}
\caption{{ \bf Pipeline.} { \it (a) Given an uncurated Internet collection (i), our framework creates a filtered collection (iii). This step is performed in a self-supervised manner using a self-filtering approach (ii): a StyleGAN generator along with an image encoder are trained on the unfiltered data; the training images are then filtered according to their reconstruction quality (measured by LPIPS). Since the generator focuses on objects with high prevalence, rather than eccentricities, we achieve a filtered collection that is more suitable for training a StyleGAN generator. (b) At inference time, we employ a multi-modal variant of the commonly used ``truncation trick'': instead of pushing the generated images towards a global average, we first cluster the latent space and then push it towards the perceptually nearest cluster center. This allows us to generate diverse, high-quality images while capturing the multi-modal nature of the data.}}
\label{fig:architecture}
\end{figure*}

%% file: related.tex
\input{figures/filtering_illustration}

\section{Related Work}

\paragraph{StyleGAN-based synthesis and editing}
A unique property of StyleGAN~\cite{karras2019style, karras2020analyzing, Karras2020ada}, is its semantically rich and disentangled latent space, which has placed StyleGAN as a prominent model for various image generation and editing tasks. While earlier editing approaches required additional supervision~\cite{shen2020interpreting,denton2019detecting, goetschalckx2019ganalyze, abdal2020styleflow}, most recent works are completely unsupervised~\cite{harkonen2020ganspace,shen2020closedform,voynov2020unsupervised,wang2021a, wu2020stylespace, patashnik2021styleclip, xia2021tedigan,gal2021stylegannada}, potentially enabling 
the semantic editing of any domain, even for real images~\cite{roich2021pivotal}. 
However, these methods have been mostly explored over aligned and curated datasets, e.g. faces, since StyleGAN struggles to achieve its distinguished realism over more challenging uncurated datasets, such as internet photos collections.

Motivated by StyleGAN's capabilities, many methods have modified the StyleGAN architecture to improve generation quality and enable additional applications over more challenging domains. 
Sendik et al.~\shortcite{sendik2020unsupervised} replace the single learned constant with multiple constants to better represent a multi-modal data distribution. Other works~\cite{Lewis2021TryOnGANBT,albahar2021pose}
incorporate pose conditioning to the generator
and perform various editing procedures over full-body fashion images.
Casanova et al.~\shortcite{casanova2021instanceconditioned} train their generator conditioned on a single instance, demonstrating the generation of novel images that share high similarity to the conditioning sample.
Their work is based on the observation that conditional GANs better reproduce complex data distributions~\cite{luvcic2019high,liu2020diverse} compared to unconditional models. 
However, these conditional methods fail to produce the desired latent semantic editing. 
In contrast, our goal is to take \textit{unconditional} StyleGAN into the uncurated Internet collections. We keep the StyleGAN architecture intact, and instead take a self-filtering approach, employing the generator itself to better handle the challenging datasets.

Karras et al.~\shortcite{karras2021alias} have suggested several enhancements (a.k.a StyleGAN3) to better address existing aliasing problems, leading to considerably smoother interpolations. 
However, it is yet to be shown that the semantic disentanglement is still preserved. Therefore, we base our framework on the irrefutable StyleGAN2 model. See ablation study in Appendix \ref{sec:appendix_ablation} for further discussion.

\vspace{-0.1cm}
\noindent
\paragraph{Dataset Distillation}

Many attempts have been made to improve the quality of large image collections. Usually, manual annotations are exploited to train a discriminative model, capable of labeling the data or filtering out the outliers. For instance, Yu et al.~\shortcite{yu2016lsun} manually label a small set, later used to train a classifier. Images that remain ambiguous to the classifier are labeled to further refine the classifier. 
\cite{KrambergerPotocnik2020} employ a detection network to filter car images containing other objects besides a car and a driver, and align the cars based on the bounding box to improve StyleGAN's generation quality.
Contrarily, our proposed method does not require any supervision.
Most close to ours is the approach suggested by Averbuch-Elor et al.~\shortcite{averbuch2015distilled}, producing a clean and structured subset of inliers for data-driven applications out of a collection of internet images, without any supervision. 
Nevertheless, 
their distillation scheme is not ideal for StyleGAN, as it intentionally ignores the background which is an inherent part of the synthesis.

\noindent
\paragraph{Latent Space Truncation}
The truncation trick~\cite{marchesi2017megapixel} became widely used as part of the BigGAN~\cite{brock2018large} architecture. 
It has been shown that truncating the initial noise vector by resampling the values with magnitude above a chosen threshold leads to a substantial quality improvement. As BigGAN generation utilizes not only a noise vector but also an auxiliary class condition, the multi-modal nature is preserved. Karras et al.~\shortcite{karras2019style} have proposed a StyleGAN variant, where the sampled latent code is interpolated with the mean latent code, showing 
a similar improvement.
Kynkäänniemi et al.~\shortcite{kynkaanniemi2019improved} have studied several truncation methods for StyleGAN. They conclude that the two preferable methods are clamping the latent vectors to the boundary of the higher density region and the proposed StyleGAN truncation. However, as we demonstrate in this paper, these truncation methods are not adequate for multi-modal data.

%% file: figures/filtering_illustration.tex
\begin{figure*}
\vspace*{-0.2cm}
\centering
{\includegraphics[width=\textwidth]{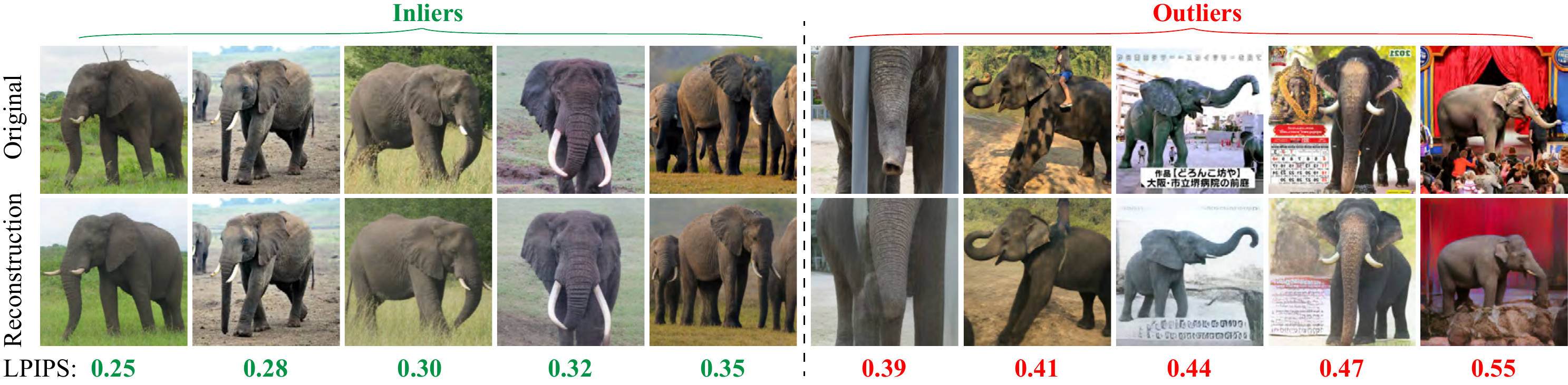}}
\vspace*{-0.6cm}
\caption{{{\bf Self-filtering examples for Internet Elephants.} \it{Top to bottom: original images, their reconstruction using our encoder-generator model, and the corresponding LPIPS distances between the images. We automatically detect outliers using the reconstruction quality: as can be seen, the reconstruction of the outliers (right) is substantially inferior compared to the inliers (left).
We visually observed that $\theta_{\text{LPIPS}}= 0.36$ typically indicates reasonable image reconstruction quality. 
}}}
 \afterfigure

\label{fig:distill_thresh}
\end{figure*}

%% file: method.tex
\section{Generative-Based Self-Filtering}
\label{sec:filterting}

Large collections of photos can be simply gathered from the internet. Yet, these contain a large portion of outliers and noisy images. A few such examples can be seen in Fig.~\ref{fig:distill_thresh} (right), where the internet elephant photos involve sculptors, occasional humans, and text. 
Training a generative model over these unfiltered datasets results in significant artifacts; state-of-the-art generators cannot model such extreme data irregularities, resulting in degraded image synthesis as illustrated in Fig.~\ref{fig:distill_compare}.  
We thus propose an automatic \emph{self-supervised} filtering approach (referred to as "self-filtering"), which is applied to the downloaded internet images, and results in high-quality image generation for multiple challenging domains. 
Although many heuristic approaches can be proposed for filtering the data (e.g., filtering specific objects or attributes), those would not necessarily lead to improvement in the image generation process of the specific generator at hand.
Instead, our self-filtering approach is \emph{generative-based}, exploiting the \emph{inherent biases} of the specific type of generator in order to identify outliers and noisy images in the raw data.

Naturally, the generator focuses on the most prevalent visual elements in the data, rather than on the tails of the distribution~\cite{yu2020inclusive}.
We leverage this generative prior -- the ability of the generator to capture and synthesize certain visual elements -- to filter the raw training data.
Our self-filtering approach, illustrated in Fig.~\ref{fig:architecture}$(a)$, consists of two steps: (i)~First, we train the generator over the undistilled images to obtain an initial generative prior. To measure its reconstruction, we jointly train the generator with an encoder. (ii)~We then measure the perceptual reconstruction quality attained for each image in the raw unfiltered image collection, and keep only those that pass a specific threshold. These 2 steps, including the threshold selection, are described in more details below.

Given a set of $N$ images, denoted $\{x_i\}_{i=1}^{N}$, our goal is to produce a filtered set $\{y_i\}_{i=1}^{M}$. In the first step, we jointly train a generator $G$ and an encoder $E$. We use the popular StyleGAN2 generator~\cite{karras2020analyzing}, and train it to produce realistic images using the original objective as have been employed by Karras et al.~\shortcite{karras2017progressive, karras2019style, karras2020analyzing}, denoted $\mathcal{L}_{GAN}$. The encoder is trained to reconstruct both real and synthesized images, similar to the work of Lang et al.~\shortcite{lang2021explaining}. Let $G_n(z)$ be an unconditionally generated image from normally distributed noise vector $z$, we denote its reconstruction as $G(E(G_n(z)))$, where $G(w)$ refers to applying the generator over latent code $w$ (i.e. skipping the mapping network). The reconstruction loss is then defined as:
\begin{align*}
\mathcal{L}_{rec,\text{LPIPS}} &= \norm{x- G(E(x))}_\text{LPIPS} + \norm{G_n(z) - G(E(G_n(z)))}_\text{LPIPS} ,\\
\mathcal{L}_{rec,L1} &= \norm{x- G(E(x))}_{L1} + \norm{G_n(z) - G(E(G_n(z)))}_{L1} ,
\end{align*}
and the reconstruction of the latent encoding:
\begin{align*}
\mathcal{L}_{w} = \norm{w - E(G(w))}_{L1} .
\end{align*}
Overall, we simultaneously train both the generator and the encoder using the final objective:
\begin{align*}
\mathcal{L} = \mathcal{L}_{GAN} + \lambda_{w}\mathcal{L}_{w} + \lambda_{L1}\mathcal{L}_{rec,L1} + \lambda_{\text{LPIPS}}\mathcal{L}_{rec,\text{LPIPS}} .
\end{align*}
$\text{LPIPS}$ is the perceptual loss presented by Zhang et al. \shortcite{zhang2018unreasonable}, $L1$ is a pixel-wise $l1$ norm, and $\lambda_{w},\lambda_{L1},\lambda_{\text{LPIPS}}$ are positive constants carrying the same values for all experiments.

\input{figures/truncation_grid}
\input{figures/truncation_single_image}

After a rather quick training procedure, we employ $G$ and $E$ to measure the $\text{LPIPS}$ reconstruction of the images $\{x_i\}_{i=1}^{N}$:
\begin{align*}
d_{\text{LPIPS}}(x_i) = \norm{x_i- G(E(x_i))}_\text{LPIPS} .
\end{align*}
The self-filtered set $Y=\{y_j\}_{j=1}^{M}$ contains all the images whose measures are bellow the threshold $\theta_{\text{LPIPS}}$: 
\begin{align*}
Y = \{x_i \: | \ \: (d_{\text{LPIPS}}(x_i) < \theta_{\text{LPIPS}})  \ \}_{i=1}^{N}  .
\end{align*}

Naturally, a trade-off emerges between 
the \emph{lower diversity/fidelity} of the filtered subset $Y=\{y_j\}_{j=1}^{M}$ with respect to the original raw set of images $X=\{x_i\}_{i=1}^{N}$, and the \emph{higher visual quality/realism} of the generated images attained by training a generator over the filtered subset $Y$. 
We therefore choose our self-filtering threshold in a way that tries to \emph{optimize the trade-off between those factors} -- i.e., strive for higher visual generation quality, while trying to minimize the loss in fidelity/diversity compared to the raw dataset. This is achieved by monitoring the increase in the FID score~\cite{heusel2017gans} between the filtered subset $Y$ and the original set $X$, as a function of the LPIPS threshold $\theta_{\text{LPIPS}}$: As long as the loss in diversity (increase in $FID(Y,X)$) is not substantial, we can continue decreasing the LPIPS filtering threshold $\theta_{\text{LPIPS}}$.

More specifically, we visually observed that $d_{\text{LPIPS}}=0.36$ typically indicates reasonable image reconstruction quality. We therefore set our threshold to be $\theta_{\text{LPIPS}}=0.36$. However, we use two additional important constraints to bound our filtering process: (i)~In order to prevent unreasonable loss in diversity due to over-filtering,
we do not allow $FID(Y,X)$ to exceed an \emph{upper-bound} $FID(G(E(X)),X)$.
(ii)~To guarantee good visual generation quality and generalization capabilities, we do not allow for the size of $Y$ to drop below $70$K training images (inspired by the size of FFHQ~\cite{karras2019style}).
$FID(G(E(X)),X)$ provides a \emph{quick and rough} measure for the diversity attainable when training a StyleGAN generator on the original \emph{unfiltered} dataset $X$. Due to the many distracting outliers in the dataset, the generation quality and reconstruction capabilities of such an initial generator $G$ are quite low. This is captured by measuring the initial fidelity loss (FID score) between the reconstructed set $G(E(X))$ and the original set $X$. This serve as a \emph{dataset-specific upper-bound} on the loss of diversity that we allow to introduce while filtering out images.

Once the self-filtering is finalized, we train the final generator model, from scratch, on the self-filtered set $Y=\{y_i\}_{i=1}^{M}$ using the native training scheme of StyleGAN2 (i.e. without an encoder). To reduce the running time overhead, we perform the self-filtering step with lower resolution images (e.g. $256 \times 256$), and train the final model with higher resolution images (e.g. $1024 \times 1024$).

\section{Multi-modal Truncation}
\label{sec:multimodal}

\input{figures/distillation_compare_lines}

The filtering process results in more coherent and clean data. However, Internet images tend to be rich and diverse, consisting of multiple modalities which constitute different geometry and texture characteristics. For example, a dataset of Lions presents different poses, zooms, and gender. A dataset of Parrots presents uniquely different  compositions of colors and feather arrangements. 
Such large diversity and data richness are of course desirable. 
We observe that the StyleGAN2 generator itself is capable of producing multi-modal samples. However, the commonly used ``truncation trick'' is not adequate for the different modalities. In particular, it often generates invalid \emph{mixtures of modalities}, e.g. parrots with invalid non-existent color combinations, distorted poses of lions, etc.

The ``truncation trick'' is employed to produce more visually realistic results, by avoiding the use of latent codes residing in the distant and sparse margins of the distribution. For this purpose, Karras et al.~\shortcite{karras2019style, karras2020analyzing} proposed the following StyleGAN variation. The sampled latent code $w$ is interpolated with the mean latent vector $\overline{w}$, referred to as the global mean, using a predefined parameter $\psi$. More formally, the truncated code is defined as $w_t = \psi \cdot w + (1-\psi) \cdot \overline{w}$. This truncation has been demonstrated to be useful for the showcasing of uncurated results, 
trading diversity for better quality and realism. However, we find the degradation of the data diversity more acute when the dataset is multi-modal. In this case, mode collapse emerges, as all samples are pushed toward the specific single modality of the global mean. 
For instance, truncating samples to the global
mean induces greenish feathers in the Parrots domain (Fig.~\ref{fig:truncation_grid}b), and a more uniform pose in the Dogs domain (Fig.~\ref{fig:truncation_single_image}c).

To this end, we introduce a \emph{multi-modal truncation} variant, enabling us to maintain the diversity of the generated images while still improving their visual quality. Our key idea is to incorporate multiple cluster centers, each representing a different modality. A sampled latent code is then truncated towards the most similar center, which results in preserving the unique perceptual attributes shared across this modality. As the cluster centers correspond to dense areas within the latent space, we obtain highly realistic results. More specifically, our multi-modal approach consists of two parts (which take place \emph{after} training a StyleGAN2 on our self-filtered data):  First,  we cluster $60,000$ randomly sampled latent codes $\{w_j\}_{j=1}^{60K}$ into $N$ clusters, obtaining $N$ cluster centers $\{c_i\}_{i=1}^N$. This step is performed only once.
Then, when a new images is generated, we assign its latent code $w$ to the ``nearest''  cluster center $c_i$, to produce its final truncated code $w_t = \psi \cdot w + (1-\psi) \cdot c_i$.

Our implementation employs the standard KMeans algorithm, where we sample the latent codes by simply passing random noise vectors through StyleGAN's mapping network~\cite{karras2019style}. We assign the ``nearest'' cluster center using the \emph{LPIPS perceptual distance}~\cite{zhang2018unreasonable}, as we find it slightly more stable and visually coherent than euclidean distance in the latent space.
The number of clusters in our experiments was set to be $N=64$ for all datasets.
We found our method to be insensitive to the exact number of clusters.
For a justification and ablation study on the number of cluster, see Appendix \ref{sec:appendix_ablation}. 
An illustration of the multi-modal truncation process is provided in Fig.~\ref{fig:architecture}$(b)$.

As shown in Figures \ref{fig:truncation_grid} and \ref{fig:truncation_single_image}, our multi-modal truncation yields results more faithful to the original generated image before truncation. 
This provides \emph{higher data diversity} compared to truncation to the global mean, \emph{while preserving high visual quality} (see Sec.~\ref{sec:experiments}).

\input{figures/truncation_fid_graph}

%% file: figures/truncation_grid.tex
\begin{figure}
\includegraphics[width=\columnwidth]{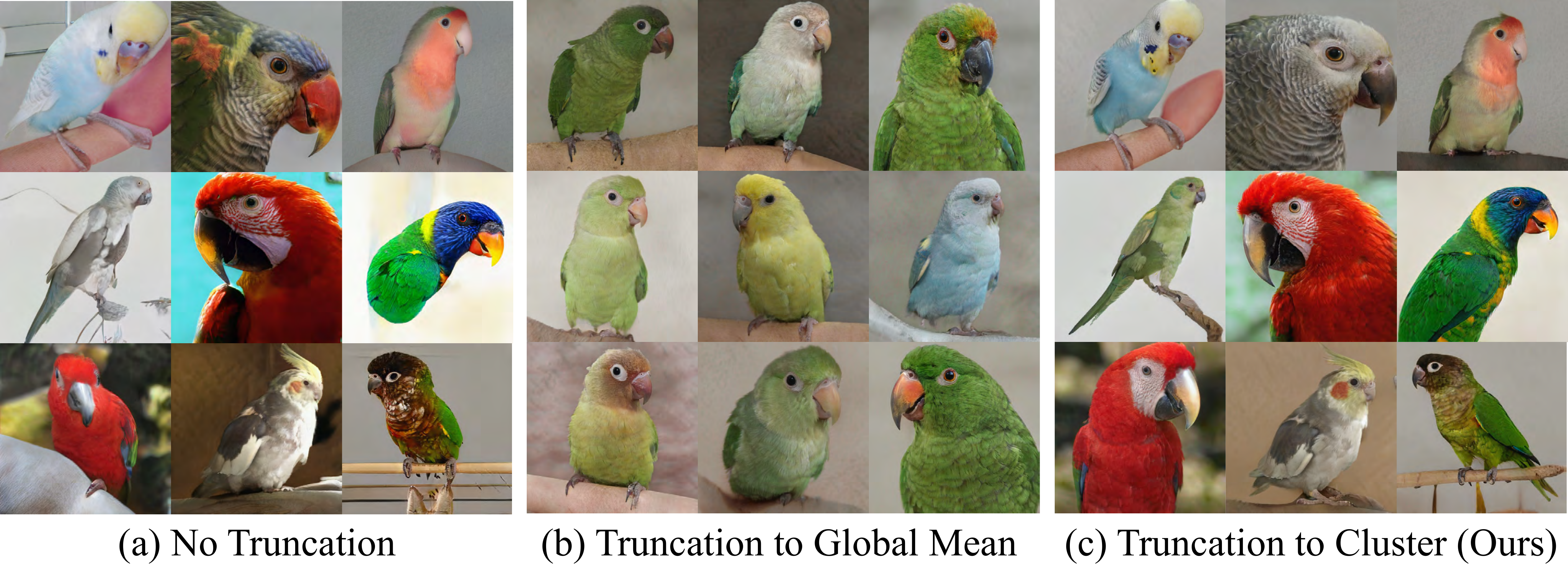}
 \vspace{-0.7cm}
\caption{
{\bf Diversity preservation of multi-modal truncation.} When sampling images (a) without applying any truncation, the generated images present different poses and color compositions. When truncated to the global mean (b), the same samples exhibit greenish feathers and a more uniform pose. Our multi-modal truncation (c) preserves better the diversity of the generated image while still improving their visual quality.
}
\afterfigure
\label{fig:truncation_grid}
\end{figure}

%% file: figures/truncation_single_image.tex
\begin{figure}
{\includegraphics[width=\columnwidth]{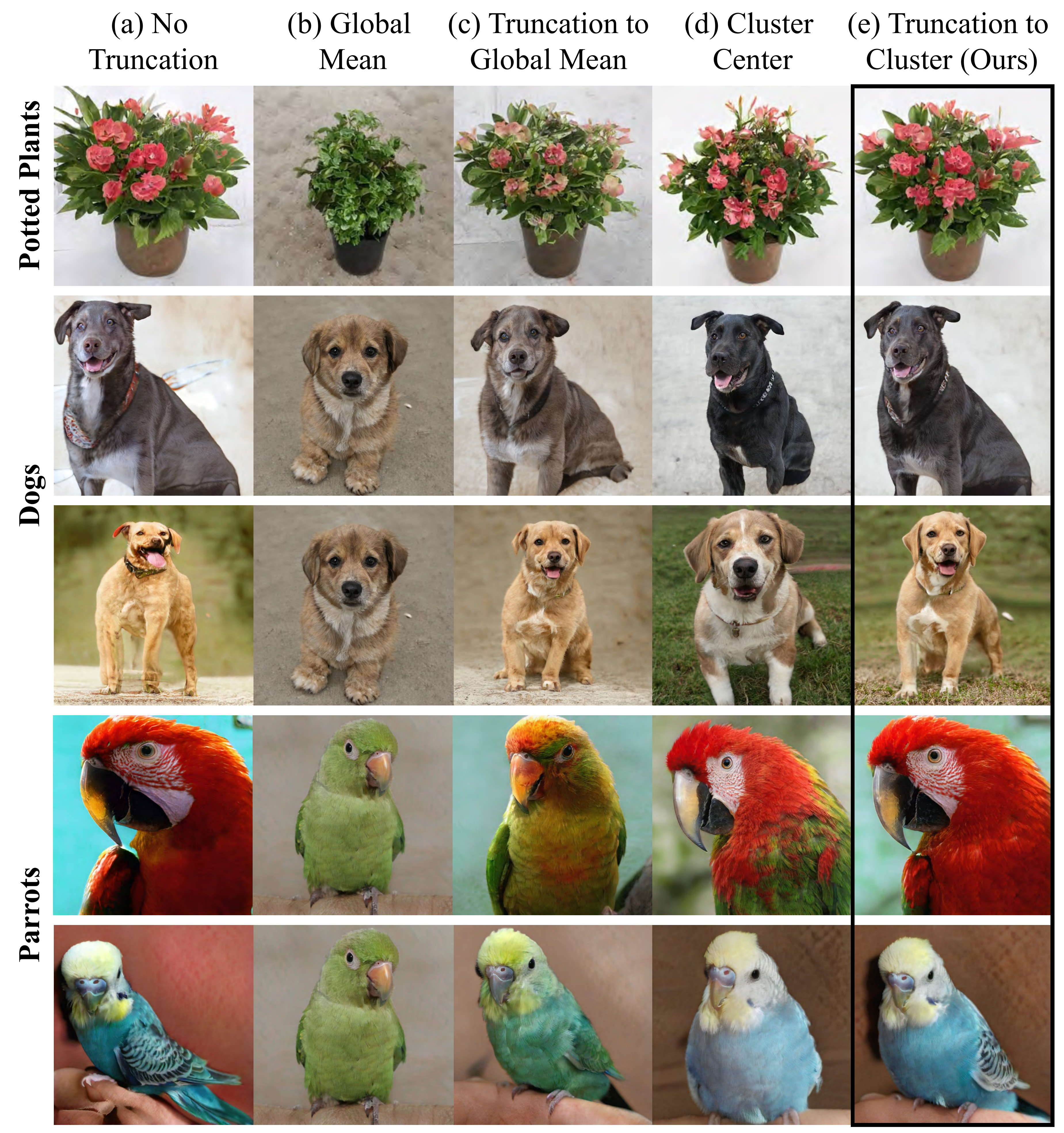}}
 \vspace{-0.7cm}
\caption{{{\bf Global mean vs. multi-modal truncation.} \it{Sampled images (a) are truncated to the global mean (b). The resulting images (c) depict low fidelity to the original image and dominate visual attributes are lost. By associating each sample with a perceptually similar cluster (d), our truncated images (e) better preserve the original image with improved quality.
}}}
\afterfigure
\vspace*{-0.2cm}
\label{fig:truncation_single_image}
\end{figure}

%% file: figures/distillation_compare_lines.tex
\begin{figure}
\centering

{\includegraphics[width=\columnwidth]{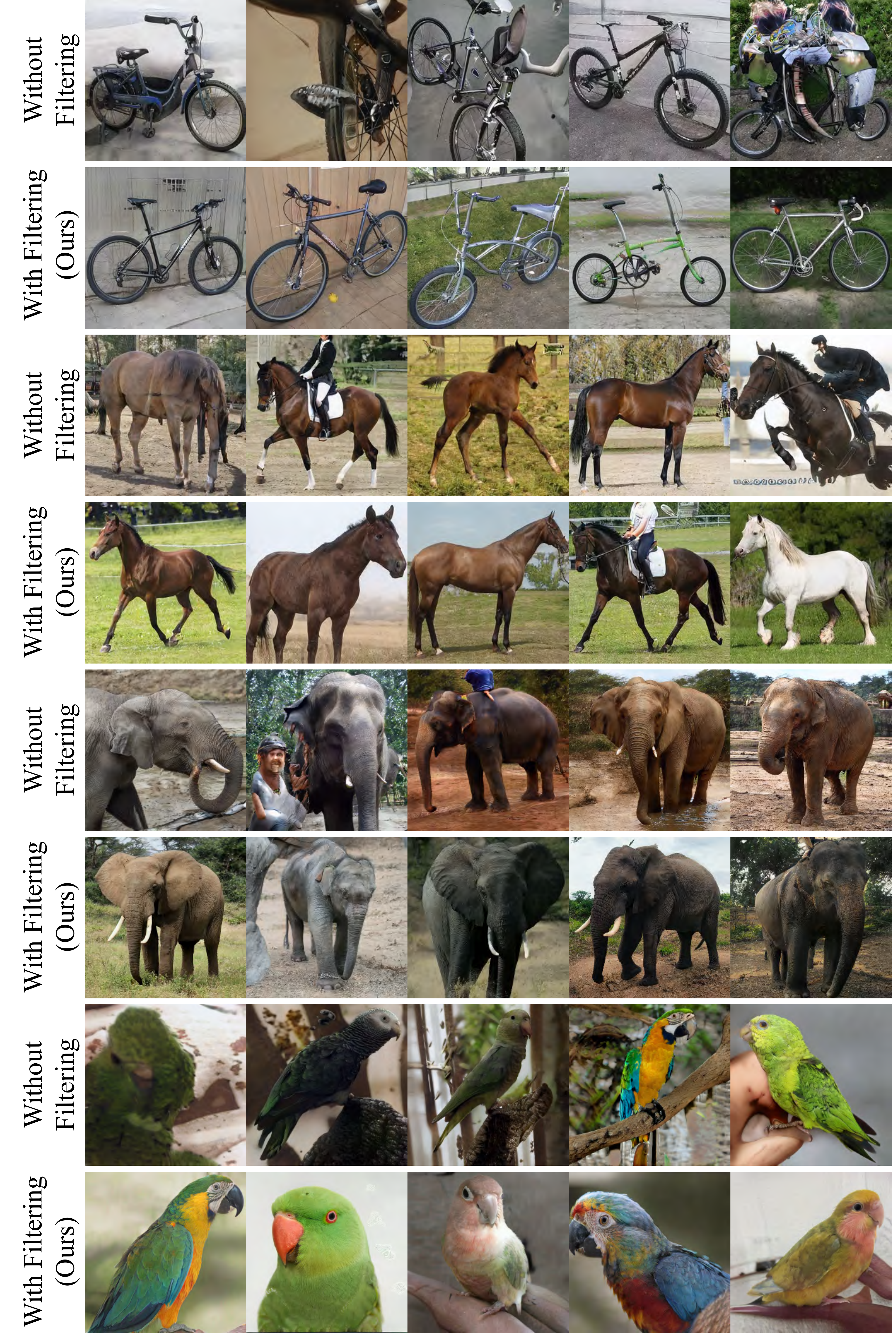}}
\vspace{-0.65cm}
\caption{{\bf Generation w/ and w/o self-filtering.} { \it Random generated images using two generators: one trained over the unfiltered collection (\textit{``Without Filtering''}), and the other trained on our self-filtered data (\textit{``Ours''}), on different domains. From top to bottom: bicycles, horses, elephants, and parrots. As shown, unrealistic artifacts emerge w/o filtering.}}
\label{fig:distill_compare}
\vspace*{-0.3cm}
\end{figure}

%% file: figures/truncation_fid_graph.tex
\begin{figure}
 {\includegraphics[width=\columnwidth]{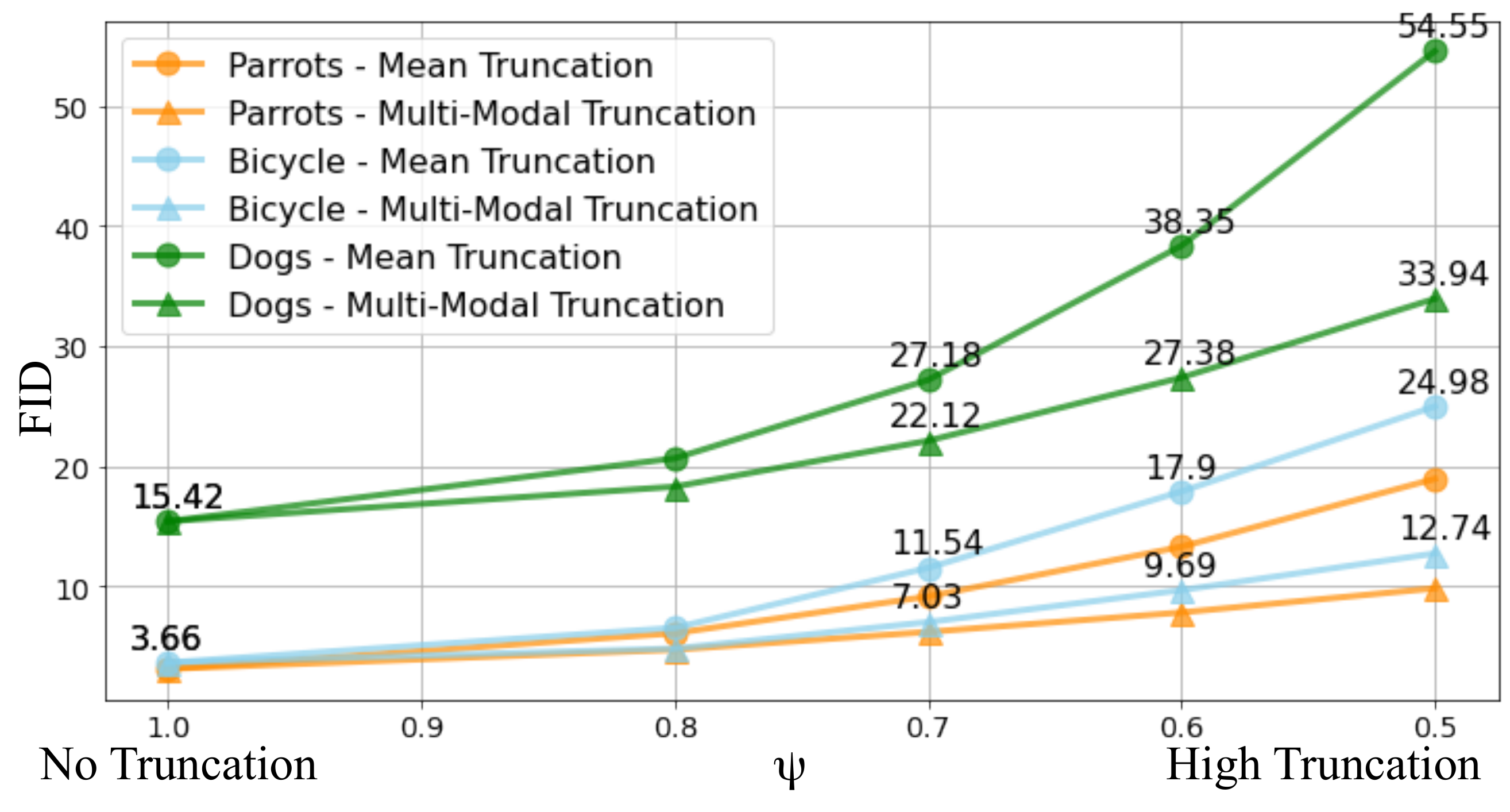}}
 \vspace{-0.7cm}
\caption{{\bf FID as a function of truncation level.} { \it FID scores measured for different truncation levels ($\psi$) over different datasets (marked in different colors) for global mean truncation (circle marker) and for our multi-modal truncation (triangle marker). For both truncation methods, FID increases with the level of truncation,  demonstrating its harm to the diversity of the data.  Our multi-modal truncation consistently results in better FID scores for the same truncation levels, and this gap increases with the level of truncation. }}
\label{fig:fid_truncation_graph}
\afterfigure
\end{figure}

%% file: results.tex
\section{Experiments}
\label{sec:experiments}

\paragraph{Internet Datasets.}
We tested our framework on several large-scale image collections taken from the LSUN~\cite{yu15lsun} dataset, as well as our own new Internet collections. For LSUN, we use \textbf{bicycles}, \textbf{potted-plants}, and \textbf{horses} domains as we found them extremely challenging, resulting in significant artifacts when applied without our method (see Fig.~\ref{fig:distill_compare}). 
However, since LSUN contains mostly low-resolution images, we gathered additional high-resolution collections crawled from the Web for challenging domains --- \textbf{lions}, \textbf{elephants}, \textbf{dogs}, and \textbf{parrots}, denoted as \textit{Internet} datasets. More specifically, we trained the generator on the LSUN dataset with resolution of $256 \times 256$, while for internet datasets we used resolution of $512 \times 512$, except for elephants ($1024 \times 1024$). We also present visual results of $1024 \times 1024$ dogs in the project website results page. We used a simple web crawler and gathered all images according to a simple keyword, e.g. ``parrot'' or ``dog''. In addition, for both LSUN and Internet datasets, we applied an off-the-shelf object detector to automatically crop the photos around the main object, after which each dataset contains at least $150,000$ images.
This allows us to eliminate irrelevant samples such as images in which the target object covers only a small region. Nevertheless, after this step, the datasets still contain a large portion of outliers including images containing cluttered backgrounds, rare objects e.g., people, cartoons, unnatural scenes or text. We demonstrate our method is not limited to cropped data by providing results over the uncropped LSUN-horse in the project results page. 
All of our \textit{Self-Distilled Internet Photos} (SDIP) datasets that are not proprietary were made publicly available. For this purpose, we have created additional sets by gathering Flickr images with the appropriate licenses. These are named \textit{SD-Flickr}. In addition, we publish the cropped and filtered LSUN datasets, named \textit{SD-LSUN}. Both are available in our  \href{https://self-distilled-stylegan.github.io/}{project website}.

\input{tables/filtering-fid_and_user_study}

\subsection{Self-Filtering Evaluation}
\label{sec:filteringEvaluation}

\input{figures/fid_graph}

To evaluate the effect of our self-filtering framework, we compare the generation results of two StyleGAN models: one trained on raw unfiltered Internet collections, and the  other trained on the self-filtered subset of the data (see Sec.~\ref{sec:filterting}). Fig.~\ref{fig:distill_compare} shows random generation results for different domains. As can be seen, unrealistic artifacts emerge when employing the unfiltered dataset (e.g., rightmost example in \emph{Bicycles}). It is also apparent that the generator struggles with generating objects, which are rare in those datasets, such as humans, leading to unreal blurry figures (e.g., second left example in \emph{Elephants}).

Our self-filtering framework relies on the reconstruction quality of our initial generator and encoder trained on the unfiltered dataset, which we assess using a perceptual similarity (LPIPS). However, our final generator is trained without an encoder on the filtered subset of data. We thus evaluate its quality by measuring FID scores of the \emph{generated images w.r.t. the training subset}.  Table~\ref{tab:filtering} reports these FID scores (lower is better) for a variety of datasets, using $50K$ randomly sampled images. As seen, the FID scores are consistently lower across datasets when self-filtering is applied. This validates the premise that StyleGAN performs better for cleaner and more structured data.

As discussed in Sec.~\ref{sec:filterting}, there is an inherent tradeoff between quality and diversity, as also illustrated in Fig.~\ref{fig:fid_graph}: reducing the LPIPS filtering threshold decreases the diversity of the filtered set $Y$ w.r.t. the raw unfiltered collection $X$ (measured by $FID(Y,X)$). Thus, overly filtering the data might unnecessarily reduce our data diversity. To avoid over-filtering, we select a threshold that achieves high reconstruction quality (ideally, $\theta_{\text{LPIPS}}=0.36$ ), yet also maintains high diversity ($FID(Y,X) < FID(G(E(X)),X)$) and large enough training set (no less than $70K$ training images). See more details in Sec.~\ref{sec:filterting}.  These criteria are shown for two datasets in Fig.~\ref{fig:fid_graph}.  In the Elephants dataset (right), the lower bound of $70$K images halted the self-filtering process ($\theta_{\text{LPIPS}}=0.38$), whereas in the Horses dataset (left),  the diversity bound $FID(G(E(X)),X)$ halted the self-filtering process. In all other datasets, the LPIPS threshold halted the filtering.

 Furthermore, we evaluate the realism and visual quality of our results by performing an extensive user study using Amazon Mechanical Turk (AMT).  We follow a typical protocol used for evaluating the overall realism and quality of images generated in an unconditional manner~\cite{shaham2019singan, granot2021drop}: a single image is displayed for $1.5$ seconds, after which the participant decides whether the image is real or fake. The results are reported in Table~\ref{tab:filtering} rightmost column. As shown, self-filtering significantly improves the scores achieved in the realism user study, reaching $60\%-70\%$ for all the tested datasets.
 
Finally, to remove the effect of the dataset size from the evaluation of our self-filtering method, we compare it to  a generator trained on a \emph{randomly sampled subset} of the data of the same size as the self-filtered sets.  Table~\ref{tab:filtering_ablation} reports the FID scores for each generator w.r.t. its training set. As can be seen, the performance of the random baseline is similar to a generator trained on the raw unfiltered data, while our self-filtering achieves substantial improvement. This demonstrates that the choice of filtering method is significantly more substantial than the dataset size.  

\subsection{Multi-modal Truncation Evaluation}
\label{sec:truncation_evaluation}

\input{figures/editing}

\input{tables/filtering_ablation_new}

Figures \ref{fig:truncation_grid} and \ref{fig:truncation_single_image} demonstrate the effect of our multi-modal truncation (Sec.~\ref{sec:multimodal}). 
Truncating the images to the global mean induces a similar effect for all samples: all images are steered to the same canonical pose and appearance, which diminishes the diversity of the generated images and the fidelity to the original non-truncated images. In contrast, using our multi-modal truncation, each image is steered towards the most perceptually similar center. This allows us to better preserve dominant visual attributes of the original samples while gaining an improved visual quality. For example, in Fig.~\ref{fig:truncation_single_image}, the parrots turn greenish $(c)$ when truncated to the green global mean $(b)$. Similarly, the breed and pose of the dogs are lost.  Our multi-modal truncation $(e)$ successfully produces high quality images that better preserve the original pose and appearance attributes.

Similar to self-filtering,  truncation also induces a tradeoff between quality and diversity, controllable by the truncation level (parameter $\psi$).  Intuitively, our multi-modal truncation provides a better quality/diversity tradeoff than truncating all image samples towards a single global mean. To quantitatively evaluate it, we measure FID scores across different truncation levels, for both truncation methods. As the truncation level increases (lower $\psi$), the generated images become more similar to the target center, hence the visual quality improves, while the diversity decreases. Fig.~\ref{fig:fid_truncation_graph} shows the computed scores for 3 different datasets. First, one can notice that the FID scores consistently increase with the truncation level, regardless of the applied truncation method. This suggests that the FID metric is more sensitive to diversity rather than quality. Second, as can be seen, our multi-modal truncation consistently results in lower (better) FID scores than truncation to the global mean, and this FID gap only increases with the level of truncation.

To evaluate the visual quality we perform another user study using AMT, comparing the image quality under different truncation methods. 
To this end, we presented side-by-side a triplet of images (in random order) produced by applying over the same sample: (i)~our multi-modal truncation, (ii)~mean truncation, and (iii)~without any truncation. The participants are asked to choose the most realistic image (with no time limit). Note that since the level of truncation ($\psi$) has no absolute meaning that can be translated from one truncation method to another, we choose the truncation level for each setting such that the truncated generated images under both settings result in the same FID (i.e., same diversity level).  As can be seen in Table~\ref{tab:fid_truncation}, for the same final FID level, our multi-modal truncated images obtain higher visual quality. As the FID score is more sensitive to diversity, we observe that our approach improves the inherent tradeoff between diversity and quality.

\subsection{Editing}
One of the main strengths of StyleGAN is its remarkable semantic editing capabilities. Fig.~\ref{fig:editing} demonstrates that our self-distilled StyleGAN preserves this quality, enabling various semantic editing effects including change of pose, and appearance.  These results are achieved by applying editing in $\mathcal{W}$ space based on~\cite{shen2020interpreting} and in Style-Space, based on~\cite{wu2020stylespace, patashnik2021styleclip}. We provide additional results as animated GIFs in the \href{https://self-distilled-stylegan.github.io/supplementary/index.html}{project website results page}, illustrating that we can successfully apply editing with the same semantic meaning over highly diverse samples, e.g. mouth opening for lions with different poses. These editing capabilities are not harmed by our self-filtering and multi-modal truncation.

%% file: tables/filtering-fid_and_user_study.tex
\begin{table}
\begin{center}

\begin{tabular}{p{0.16\columnwidth}|cc|cc}
\toprule
\multirow{2}{*}{\shortstack[l]{Dataset}} & \multicolumn{2}{c|}{FID $\downarrow$} & 
\multicolumn{2}{c}{\%Real (Human raters)$\uparrow$} \\\cline{2-5}
&  w/o filtering & Ours  & w/o filtering & Ours   \\
\toprule
LSUN-Bicycle  & $5.42$ & \textbf{3.66} & $55.0$ & \textbf{73.8} \\
\midrule
LSUN-Horse & $4.05$ & \textbf{2.81} &  $49.7$ &  \textbf{68.2}\\
\midrule
Internet Elephants & $2.25$ & \textbf{2.05}  &  $59.6$  &  \textbf{67.3}\\
\midrule
Internet Lions &  $5.15$ & \textbf{3.43} &  $50.6$ &  \textbf{62.6}\\
\midrule
Internet Dogs  &  $19.49$ & \textbf{15.42}  &  $44.4$ &  \textbf{60.1} \\
\bottomrule
\end{tabular}

\vspace*{0.1cm}
\caption{{\bf Self-filtering quantitative evaluation.}
{ \it We measure FID of the generated images w.r.t. the training set (w/ and w/o filtering). Filtering consistently improves the FID (lower is better). We further measure the realism of our generated images, via an AMT user study (see Sec.~\ref{sec:filteringEvaluation}).
Our self-filtering approach achieves superior results in all domains in both measures.}}
\label{tab:filtering} 
\vspace{-0.05cm}
\begin{tabular}{p{0.16\columnwidth}|c|c|c} \toprule
\multirow{3}{*}{Dataset} & 
\multicolumn{3}{c}{Human raters$\uparrow$ [\% of those favored]} \\\cline{2-4}
 & \multirow{2}{*}{\shortstack[c]{Without \\ Truncation}}  & \multirow{2}{*}{\shortstack[c]{Mean \\ Truncation}} & \multirow{2}{*}{\shortstack[c]{Multi-modal \\ Truncation}}  \\  
&  &  &  \\  
\toprule
Internet Parrots & $21.7\%$ $(3.14)$ & $35.2\%$ $(9.15)$ & \textbf{43.1\%} $(8.84)$ \\
\midrule
LSUN-Horse & $13.2\%$ $(2.81)$  & $31.1\%$ $(11.36)$  & \textbf{55.7\%} $(10.58)$  \\
\midrule
Internet Lions & $15.8\%$ $(3.43)$ & $39.8\%$ $(10.51)$& \textbf{44.4\%}  $(10.70)$\\
\midrule
Internet Dogs & $8.1\%$ $(15.42)$ & $37.7\%$ $(27.18)$& \textbf{54.2\%} $(27.37)$ \\
\midrule
LSUN-Bicycle & $16.0\%$ $(3.66)$& $40.0\%$ $(11.91)$ & \textbf{44.0\%} $(11.21)$ \\
\bottomrule
\end{tabular}
\vspace*{0.1cm}
\caption{{\bf Multi-modal vs. Global-mean truncation: A user study.} \it{
Using an AMT user study, 
we compare the visual quality of images produced using: our multi-modal truncation, global-mean truncation, and no truncation.
We presented side-by-side the results obtained by applying these methods on the same sample, and report the $\%$ of raters that favored each. As described in Sec.~\ref{sec:truncation_evaluation}, we select the truncation parameters $\psi$ that lead to a similar FID score  (diversity)  for both  multi-modal and global-mean truncation (FID is reported in brackets).
Our method achieves superior realism  for comparable FID, thus improving StyleGAN's inherent tradeoff between quality and diversity. }}\afterfigure
\label{tab:fid_truncation} 
\end{center}
\vspace{-0.5cm}
\end{table}

%% file: figures/fid_graph.tex
\begin{figure}[t!]
\vspace*{-0.2cm}
\includegraphics[width=\columnwidth]{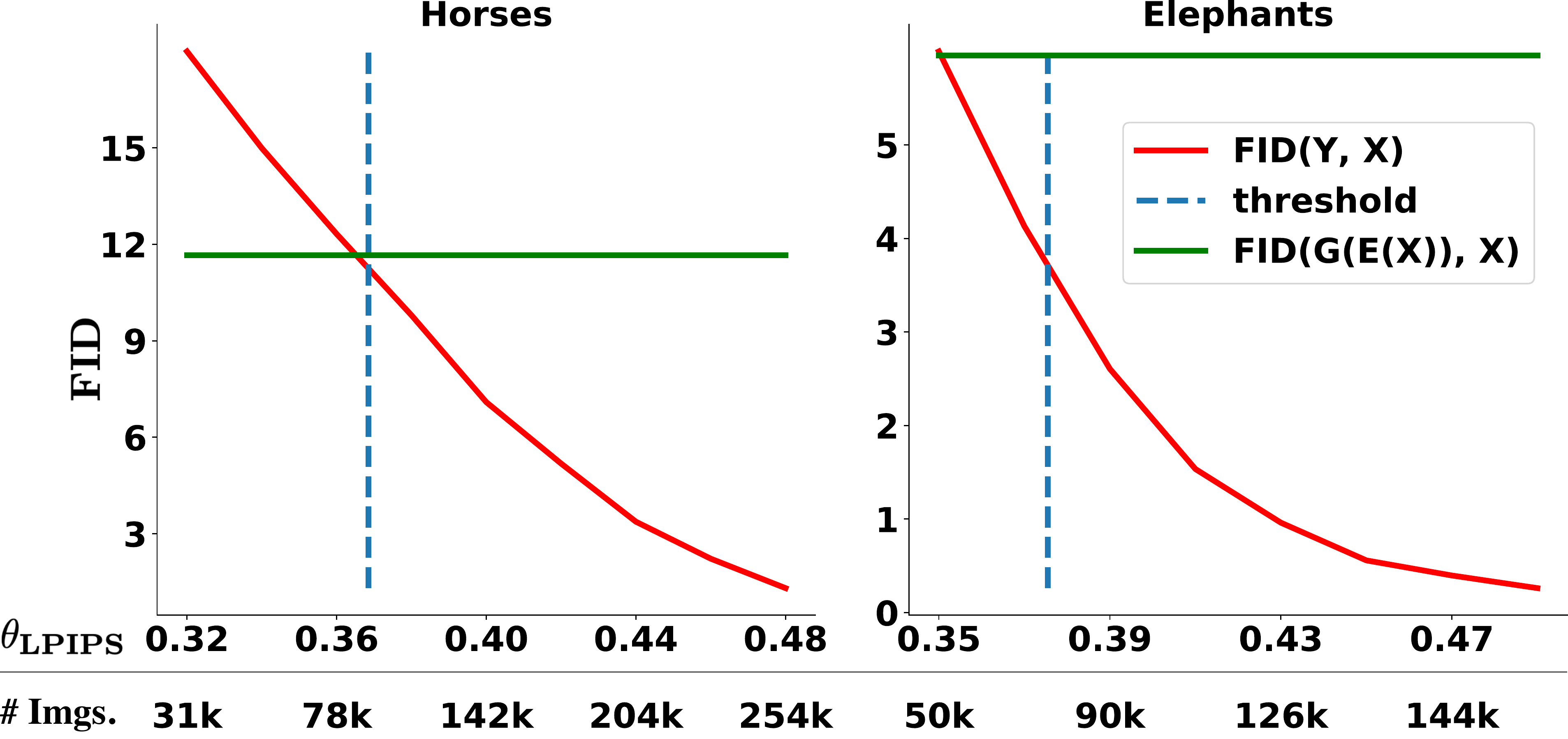}
\caption{{ \bf The diversity/quality tradeoff.} \it{We measure FID between 
the filtered subset $Y$ and the raw unfiltered data $X$,
for gradually increasing filtering threshold values $\theta_{\text{LPIPS}}$. The resulting $FID(Y,X)$ curves are shown for the \emph{Horses} and \emph{Elephants} datasets. The
$\theta_{\text{LPIPS}}$ values and the corresponding 
size of the filtered subset $Y$ are indicted 
on the x-axis.
As we lower $\theta_{\text{LPIPS}}$, we filter out more of the data, hence the mismatch between the distribution of $Y$ and $X$ increases (higher FID scores). We treat $FID(G(E(X)),X)$ -- the FID between 
the raw dataset $X$ and its reconstructions $E(X)$ 
-- as a filtering  upper-bound 
(indicated by the horizontal green line, see Sec.~\ref{sec:filterting}).  The threshold $\theta_{\text{LPIPS}}$ used in practice (indicated by the vertical dotted lines) provides a reasonable diversity/quality tradeoff.
}}
\label{fig:fid_graph}
\vspace*{-0.3cm}
\end{figure}

%% file: figures/editing.tex
\begin{figure}
\vspace*{-0.2cm}
    \includegraphics[width=0.95\columnwidth]{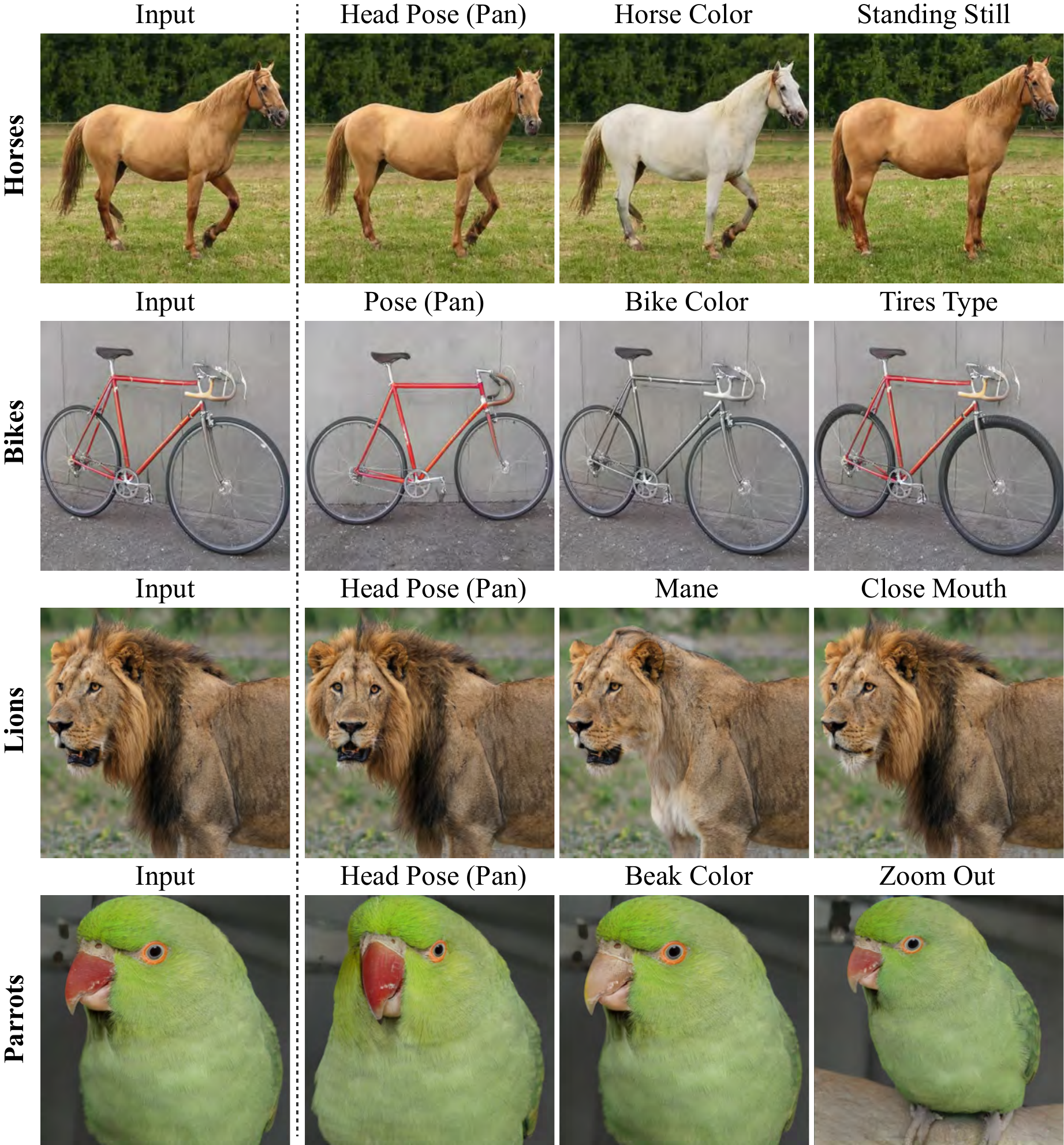}
    \vspace{-0.2cm}
    \caption{{\bf Editing.} \it{Our Self-Distilled StyleGAN retains remarkable semantic editing capabilities, enabling us to edit various attributes in each domain. Additional results, in the form of animated GIFs,  are provided in the project website results page.}}
\label{fig:editing}
\end{figure}

%% file: tables/filtering_ablation_new.tex
\begin{table}
\begin{center}


\begin{tabular}{p{0.25\columnwidth}|c|cc}
\toprule

\multirow{2}{*}{Dataset} & & \multicolumn{2}{c}{Same \#Images} \\\cline{3-4}  
 &  Unfiltered & Random & Filtered \\
\toprule
LSUN-bicycle &$5.42$ & $5.88$ & \textbf{3.66} \\
LSUN-Horse &   $4.05$ & $3.96$ & \textbf{2.81}  \\
Internet Dogs  & $19.49$ & $19.34$ & \textbf{15.42}  \\
\bottomrule
 \end{tabular}

\vspace*{0.1cm}
\caption{
{\bf Self-filtering vs. random filtering.} \it{ We compare between a generator trained on our self-filtered subset and a generator trained over \textbf{randomly sampled} subset of the raw data with the exact \textbf{same size}. The FID scores, computed for each generator w.r.t. its training set, show that the choice of filtering approach is more significant than the dataset size.}} \afterfigure
\label{tab:filtering_ablation} 
\end{center}
\vspace*{-0.7cm}
\end{table}

%% file: conclusion.tex
\section{CONCLUSIONS}

We introduced a novel approach for taking StyleGAN towards the realm of image generation from uncurated data. Our self-distilled approach leverages the generator itself in two key ways: (i)~the generator's natural bias in capturing the prevalent characteristics of the data is used to drive the self-filtering process of the raw uncurated data, and (ii)~the smooth and semantically meaningful latent space of the generator is used to capture the multi-modal distribution of the data, and to adapt StyleGAN’s ``truncation trick''  via simple clustering.   As our approach is completely unsupervised, it enables the extension of StyleGAN generation to a variety of new domains, and to new datasets that are directly gathered from the Internet.  Furthermore, by keeping StyleGAN design intact,  our self-distilled approach preserves its remarkable editing abilities. 
Nevertheless, StyleGAN's inherent tradeoff between visual quality and diversity of the generated images remains a limitation of our method (although improved).  
It still struggles to faithfully synthesize complicated backgrounds and scenes. Resolving this open challenge would likely require substantial architectural changes, which are beyond the scope of this paper. We believe that the principles we provide are a significant step towards adapting StyleGAN to more ``in the wild'' domains, and hope this work can guide and trigger further progress in that direction.

%% file: appendix.tex
\section{Implementation Details}
\paragraph{Self-filtering} 
In our experiments we use our own Tensorflow $2.0$ StyleGAN2 implementation with the same  hyperparameters as in the official implementation.
The encoder architecture is based on the StyleGAN2 discriminator. We train our encoder jointly with the StyleGAN generator using $8$ Tesla V100 GPUs for $250k$ steps, with a learning rate of $0.002$ for both encoder and generator and batch size of $16$. 
We set $\lambda_{w} = \lambda_{L1} = \lambda_{\text{LPIPS}} = 0.1$ in all our experiments. We train the generator over the filtered subset for $500K$ iterations in the comparisons and ablations and ~$1500K$ iterations for the final models.

\paragraph{Multi-Modal Truncation}
Our implementation employs the standard Kmeans algorithm, where we sample $60,000$ latent codes by passing random noise vectors through StyleGAN's mapping network. We obtain $N=64$ clusters using Python Sklearn implementation with the default parameters, except for 'init' which we set to random.

\section{Additional Ablations and Comparisons}
\label{sec:appendix_ablation}

\paragraph{StyleGAN3} 
\cite{karras2021alias}  has been demonstrated to better address unaligned data. Therefore, we validate that our filtering method is still valuable even when utilizing this architecture. Trained over the challenging LSUN-bicycle, StyleGAN3 achieves an FID score of $6.5$ on unfiltered data and $2.6$ on our filtered dataset. StyleGAN2 obtains $5.42$ and $3.66$, respectively. We conclude that StyleGAN3 alone is insufficient to handle  challenging uncurated datasets collected from the internet, and also benefits from our filtering scheme. Since it is yet to be shown that the semantic disentanglement of StyleGAN2 is still preserved in StyleGAN3, we base our framework on the irrefutable StyleGAN2 model.

\paragraph{Number of clusters.} The number of clusters, $N$, is affecting the obtained clustering quality in our multi-modal truncation. Insufficient number results in each cluster consisting of multiple modalities, leading to inferior diversity, similar to the global mean truncation. On the other hand, an excessive number of clusters yields centers with lower visual quality resulting in additional artifacts. First, We have used the commonly practiced elbow method to determine the number of clusters. We evaluated the elbow method based on KMeans inertia and FID with a fixed $\psi=0.5$. Both results in nearly $N=64$ for most datasets, while increasing $N$ to 128 clusters gains only a minor FID improvement. In addition, we study the preferable number of clusters, by conducting a user study (AMT). We presented side-by-side a triplet of images using $32$, $64$, or $128$ clusters for our proposed truncation approach. The participants were asked to choose the most realistic image. 
As can be seen in Table \ref{tab:trunc_size_comparison}, our method is not highly sensitive to the exact number of clusters. Yet, rich and diverse domains that consist of many modalities, e.g. dogs, which differ in pose and breed, achieve slightly better results with $128$ cluster centers. On the other hand, the score of less diverse domains, such as horses, slightly improves when using only $32$ clusters. Overall, we conclude that $64$ clusters perform well for all datasets, and therefore, we use this configuration for all our experiments.

\paragraph{Latent-based cluster assignment ablation.} 

We further validate our design choice, assigning the "nearest" cluster center using the LPIPS perceptual distance rather than using the euclidean distance between the latent codes. We compare the two alternatives by conducting a user study (AMT), where the participants are asked to choose the more realistic image. As shown in Table~\ref{tab:trunc_comparison}, LPIPS-based assignment indeed outperforms the latent space-based assignment.

\paragraph{Clamping truncation comparison.}
We compare our multi-modal truncation method to the clamping toward the global mean method \cite{kynkaanniemi2019improved}. Similar to other comparisons and ablations, we evaluate this ablation via an AMT user study. Again, we tune the parameters of the different methods to reach the same FID. As can be seen in Table~\ref{tab:trunc_comparison}, our results are superior, therefore, we further conclude that using clamping to global mean is also not adequate for challenging multi-modal data.

\begin{table}
\begin{center}

\begin{tabular}{c|ccc}

\toprule
 & \multicolumn{3}{|c}{Number of clusters}  \\
 Dataset &
 $32$ &
 $64$ &
 $128$ \\
 
 \toprule
Internet Lions & $34.3\%$ & $33.5\%$  & $32.2\%$ \\
Internet Parrots & $34.0\%$ & $32.0\%$ & $34.0\%$ \\
LSUN-Horses & $34.5\%$ & $34.1\%$ & $31.4\%$ \\
LSUN-Bicycles & $32.6\%$ & $32.3\%$ & $35.1\%$ \\
Internet Dogs & $26.4\%$ & $36.0\%$ & $37.6\%$ \\
\bottomrule

\end{tabular}

\end{center}
\caption{{\bf Ablation user study (AMT) results for the number of clusters used in our multi-model truncation. } {\it We presented side-by-side a triplet of images using $32$, $64$, or $128$ cluster centers for our proposed truncation scheme. The participants were asked to choose the most realistic image. Overall, we observe that $64$ clusters perform well for all datasets.
}}
\label{tab:trunc_size_comparison} 
\end{table}

\input{tables/truncation_comparison}

%% file: tables/truncation_comparison.tex
\begin{table}
\begin{center}



\begin{tabular}{p{0.17\columnwidth}ccccc}

\toprule

&Lions & Parrots  & Horses & Bicycles & Dogs \\

\toprule
 Latent \\ Assignment & $43.1\%$ & $44.5\%$ & $41.8\%$ & $44.1\%$ & $45.9\%$ \\
 Ours & \textbf{56.9\%} & \textbf{55.5\%} & \textbf{58.2\%} & \textbf{55.9\%} & \textbf{54.1\%} \\
\toprule
Clamping & $42.6\%$ & $43.5\%$ & $30.5\%$ & $45.7\%$ & $35.7\%$ \\
Ours & \textbf{57.4\%} & \textbf{56.5\%} & \textbf{69.5\%} & \textbf{54.3\%} & \textbf{64.3\%} \\

\bottomrule

\end{tabular}

\caption{{\bf Additional ablation and comparison for multi-modal truncation. }
\it{For each comparison, we conduct a user study where the raters requested to choose the more realistic result out of two given images --- ours and the evaluated baseline. We report the percentage of raters in favor of each.
Top: We replace the LPIPS-based assignment with the latent-based assignment. Bottom: we perform clamping to the global mean \cite{kynkaanniemi2019improved}.
}}
\label{tab:trunc_comparison} 

\end{center}
\end{table}